\documentclass[10pt,twocolumn,letterpaper]{article}

\usepackage{iccv}
\usepackage[accsupp]{axessibility}
\usepackage{times}
\usepackage{epsfig}
\usepackage{graphicx}
\usepackage{amsmath}
\usepackage{amssymb}

\usepackage{multirow}
\usepackage{booktabs}
\usepackage{adjustbox}

\def\argmin{\operatornamewithlimits{arg\,min}}
\newcommand{\norm}[1]{\left\lVert#1\right\rVert}

\usepackage[pagebackref=true,breaklinks=true,letterpaper=true,colorlinks,bookmarks=false]{hyperref}

\iccvfinalcopy 

\def\iccvPaperID{10705} 

\ificcvfinal\fi

\begin{document}

\title{Domain Generalization via Gradient Surgery}

\author{Lucas Mansilla \qquad Rodrigo Echeveste \qquad Diego H. Milone \qquad Enzo Ferrante\\ \\
\textit{Research Institute for Signals, Systems and Computational Intelligence,} sinc($i$)\\ \textit{CONICET, UNL, Santa Fe, Argentina}\\
{\tt\small \{lmansilla, recheveste, dmilone, eferrante\}@sinc.unl.edu.ar}}

\maketitle
\ificcvfinal\thispagestyle{empty}\fi

\begin{abstract}
In real-life applications, machine learning models often face scenarios where there is a change in data distribution between training and test domains. When the aim is to make predictions on distributions different from those seen at training, we incur in a domain generalization problem. Methods to address this issue learn a model using data from multiple source domains, and then apply this model to the unseen target domain. 
Our hypothesis is that when training with multiple domains, conflicting gradients within each mini-batch contain information specific to the individual domains which is irrelevant to the others, including the test domain. If left untouched, such disagreement may degrade generalization performance. In this work, we characterize the conflicting gradients emerging in domain shift scenarios and devise novel gradient agreement strategies based on gradient surgery to alleviate their effect. We validate our approach in image classification tasks with three multi-domain datasets, showing the value of the proposed agreement strategy in enhancing the generalization capability of deep learning models in domain shift scenarios.
\end{abstract}
\section{Introduction}

Deep learning models have shown remarkable results in diverse application areas such as image understanding \cite{krizhevsky2012imagenet, tompson2014joint}, speech recognition \cite{hinton2012deep, mikolov2011strategies} and natural language processing \cite{sarikaya2014application, sutskever2014sequence}. Such models are typically trained under the standard supervised learning paradigm, assuming that training and test data come from the same distribution. However, in real life, training and test conditions may differ by several factors, such as a change in data acquisition device or target population. This makes models perform poorly when applied to test data whose distribution differs from the training data and, therefore, limits their implementation in such real scenarios. The goal is then to develop deep learning models that generalize outside the training distribution, under domain shift conditions.
\begin{figure}
\begin{center}
    \includegraphics[width=0.8\columnwidth]{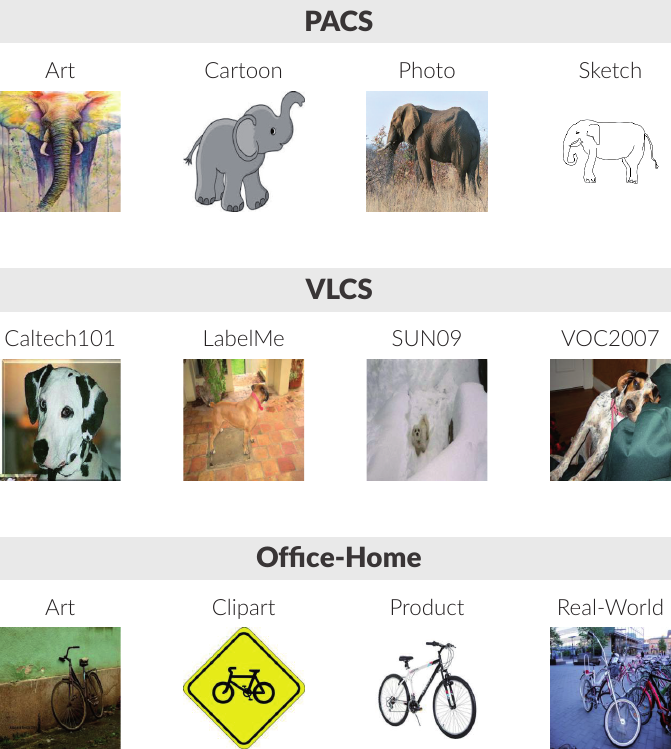}
\end{center}
   \caption{Example images extracted from three multi-domain datasets: PACS \cite{li2017deeper}, VLCS \cite{fang2013unbiased} and Office-Home \cite{venkateswara2017deep}. The goal of domain generalization is to train a model that performs well on data sampled from domains different from those seen during training.}
\label{fig:datasets}
\end{figure}

Learning a model with data from different domains and then applying it to a new domain not seen during training entails a domain generalization (DG) problem \cite{gulrajani2020search}. In the DG literature, training domains are often called source domains, while the test domain is referred as target. The problem itself is highly challenging, since not even unlabeled data from the target domain is accessible during training. Thus, the model must be trained without information about the target domain. In the particular case of image classification, for example, different domains may differ in their visual characteristics, e.g. photographic images or more abstract representations, such as paintings and sketches (see Figure \ref{fig:datasets} for visual examples). In this scenario, the main challenge is how to guide the learning process in order to capture information that is relevant to the task, and invariant to domain changes.

To address the challenges inherent to DG, different strategies have been developed over time. Proposed works have mainly focused on: i) training and fusing multiple domain-specific models \cite{xu2014exploiting, mancini2018best}, ii) learning and extracting common knowledge from multiple source domains, such as domain-invariant representations \cite{muandet2013domain, ghifary2015domain, li2018domain} or domain-agnostic models \cite{khosla2012undoing, li2017deeper, li2018learning}, and iii) increasing the data space through data augmentation \cite{shankar2018generalizing, carlucci2019domain, volpi2018generalizing}. More recently, important contributions have been made regarding model selection in the presence of domain shift \cite{gulrajani2020search}, ignored in most previous works. Albeit the great efforts made by the machine learning and computer vision communities, the gains in performance obtained by current domain generalization techniques are still modest \cite{carlucci2019domain,dou2019domain}. Thus, further research is still necessary to better understand the reasons behind this phenomenon.

In contrast to previous approaches, in this work we are specifically interested in understanding the implications of multi-domain gradient interference in domain generalization. The recent work of \cite{yu2020gradient} analyzes this problem in the context of multi-task learning (MTL) \cite{caruana1997multitask}. The authors find that one of the main optimization issues in MTL arises from gradients from different tasks conflicting with one another, in a way that is detrimental to making progress. The main hypothesis of our work is that multiple domains also give place to conflicting gradients, which are associated with different domains instead of tasks. We characterize the conflicting gradients emerging in domain shift scenarios and devise novel gradient agreement strategies based on gradient surgery to alleviate their effect. 

The gradient surgery framework was introduced in \cite{yu2020gradient} to address multi-task learning, and is rooted in a simple and intuitive idea. In general, deep neural networks are trained using gradient descent, where gradients guide the optimization process through a loss landscape. This landscape is defined by the loss function and the training data. In MTL, a different loss function is employed for each task. This can lead to conflicting gradients, i.e. gradients which may point in opposite directions when associated to different tasks. The usual way to deal with conflicting gradients is to just average them. However, the works of \cite{yu2020gradient, lopez2017gradient} recently showed that simply averaging them can lead to significantly degraded performance. Unlike MTL, in domain generalization the task remains fixed but we must handle different domains. Here we hypothesize that similar conflicts emerge when training with multiple domains. In this case, conflicting gradients within each mini-batch contain information specific to the individual train domains, which is irrelevant to the test domains and, if left untouched, will degrade generalization performance. Thus, we aim to distil domain invariant information by updating the neural weights in directions which encourage gradient agreement among the source domains. Extensive evaluation in image classification tasks with three multi-domain datasets demonstrate the value of our agreement strategy in enhancing the generalization capacity of deep learning models under domain shift conditions.

\section{Related work}

\noindent\textbf{Domain generalization.} Since DG aims to improve model performance in scenarios where there are statistical differences between the source and target domains, it is closely related to domain adaptation (DA) \cite{wang2018deep}, where domain shifts are also addressed. However, while DA assumes we have access to (labelled or unlabelled) data samples from the target domain, DG supposes that such data samples are not available during training. Therefore, DG methods have to seek solutions to better exploit the information from multiple source domains accessible during training. The hope is that distilling knowledge common to all the source domains will lead to more robust features, potentially useful in unseen target domains.

The DG methods presented to date can be divided according to the strategy they employ to achieve generalization. One group of methods is based on the idea of training a specific classifier for each source domain and then combining them optimally by measuring the similarity between source domains and test samples \cite{xu2014exploiting, mancini2018best}. Other studies propose to reduce the gap across domains using data augmentation algorithms \cite{shankar2018generalizing, carlucci2019domain, volpi2018generalizing}. An alternative approach assumes that there is a common knowledge to all domains that can be acquired from multiple sources and transferred to new domains. Some studies exploit this idea by seeking to learn a domain-invariant representation via kernel-based models \cite{muandet2013domain}, multi-task auto-encoders \cite{ghifary2015domain} and generative adversarial networks \cite{li2018domain}. Instead of domain-invariant feature representations, other methods propose extracting domain-agnostic parameters to address generalization through max-margin linear models \cite{khosla2012undoing}, low-rank parametrized CNNs \cite{li2017deeper} and meta-learning \cite{li2018domain, dou2019domain}.\\

\noindent\textbf{Gradient surgery in the context of MTL.} MTL aims to improve generalization performance by leveraging domain-specific information from a set of related tasks \cite{caruana1997multitask}. To achieve this, MTL techniques typically train a single model jointly for all tasks by assuming that there is a shared structure across them that can be learned. In practice, training a model that can solve multiple tasks is difficult, as defining appropriate strategies for balancing and controlling multiple tasks is required.
Gradient surgery refers to a number of techniques that have been introduced to improve the learning process of MTL models by directly operating on individual task-specific gradients during optimization. Chen et al. \cite{chen2018gradnorm} introduce a gradient normalization algorithm (GradNorm) that balances the contribution of each task by scaling the magnitudes of task-specific gradients dynamically, allowing different tasks to be trained at similar rates. Yu et al. \cite{yu2020gradient} discuss the conflicting gradient problem, which arises when the gradients of different tasks point in opposite directions given by negative cosine similarity, and present PCGrad, a method to mitigate gradient conflicts. PCGrad removes the component causing interference by projecting the gradient from one task onto the normal component of the gradient from the other, mitigating the negative-cosine similarity problem. More recently, Wang et al. \cite{wang2020gradient} generalize this idea by proposing an adaptive gradient similarity method (GradVac) that allows setting an individual gradient similarity objective for each task pair to better exploit inter-task correlations.\\

\noindent\textbf{Contributions.} In this study we propose a gradient surgery strategy to tackle domain generalization problems. Inspired by previous works on muli-task learning, we characterize conflicting gradients emerging in single-task scenarios with multiple domains, instead of tasks. As expected, we show that intra-domain gradients tend to exhibit higher similarity than their inter-domain counterpart, and propose novel gradient agreement variants to encourage the learning of those discriminative features that are common to all domains. Our results suggest that updating neural weights in directions of common accord by harmonizing inter-domain gradients helps to create more robust image classifiers. Compared to standard gradient descent and existing PCGrad techniques, our agreement strategies tend to produce models with better generalization performance in unseen image domains. 
\section{Methods}

\subsection{Preliminaries for domain generalization}
In a DG setting, we have access to a training set composed of $N$ source domains $\mathcal{D}=\{D_1,D_2,...,D_N\}$, where the $i$-th domain is characterized by a dataset ${D_i=\{(x_{j}^{(i)},y_{j}^{(i)})\}_{j=1}^{M_i}}$ containing $M_i$ labeled data points, and all domains have the same number of classes. The aim is to learn a classification function  $f(x_{j}^{(i)};\theta)$ which predicts the class label $\hat{y}_{j}^{(i)}$ corresponding to the input $x_{j}^{(i)}$ with competitive performance in all the source domains, but can also generalize to unseen target domains. Here, $\theta$ denotes the model parameters to be learned. For multiple source domains, we define the training cost function as the average loss over all source domains ${\mathcal{L}(\theta)=\frac{1}{N}\sum_{i=1}^{N} \mathcal{L}_{i}(\theta)}$, where ${\mathcal{L}_{i}(\theta)=\frac{1}{M_i}\sum_{j=1}^{M_i} \ell\left(f(x_{j}^{(i)};\theta),y_{j}^{(i)}\right)}$ represents the loss associated to the $i$-th domain. The function $\ell(\cdot,\cdot)$ is a classification loss, e.g. cross-entropy, which measures the error between the predicted label $\hat{y}$ and the true label $y$. We train the model by minimizing the following objective:

\begin{equation}
    \hat{\theta} = \argmin_{\theta} \frac{1}{N}\sum_{i=1}^{N}
    \mathcal{L}_{i}(\theta) + \lambda R(\theta),
\label{eq:objective}
\end{equation}

\noindent where $R(\cdot)$ is a regularization term  included to prevent overfitting, while the parameter $\lambda$ controls its importance. After training on source domains, the final model with the learned parameters $\hat{\theta}$ is evaluated on the target domain, where samples may come from a different distribution.

\subsection{Domain generalization via gradient surgery}
The typical strategy used to train classification models with multiple source domains is to simply create mini-batches by randomly sampling from all sources with equal probability. In this context, standard mini-batch gradient descent is then used to optimize the objective function defined in Eq. \ref{eq:objective}. Following the literature on domain generalization \cite{dou2019domain,carlucci2019domain}, we refer to this approach as \emph{Deep-All}. 

Here we propose to modify the standard mini-batch gradient descent by incorporating a gradient surgery step before updating the neural weights, while optimizing the objective function defined in Eq. \ref{eq:objective}. 
The goal of our approach is to adjust model parameters $\theta$ by modifying gradient updates so that they point in a direction that improves the agreement across all domains. Such harmonization step will be defined according to the sign of the respective components of the gradient vectors associated to each domain. Intuitively, given a collection of gradient vectors (one per domain), we will construct consensus vectors by retaining those components that point in the same direction (i.e. those with the same sign) and modifying the conflicting components. Here we define two different strategies to deal with the conflicting components: we either set them to zero (we refer to this strategy as \emph{Agr-Sum}) or we assign a random value to them (we refer to this as \emph{Agr-Rand}). In what follows, we discuss the proposed approaches in detail.\\

\noindent \textbf{Agr-Sum consensus strategy. } Given a set of training source domains we first sample a mini-batch from each source. Next, we perform a forward pass through the network, and compute the domain losses $\mathcal{L}_i$ and the corresponding gradients $g^{(i)}=\nabla_{\theta} \mathcal{L}_{i}(\theta)$ via backpropagation. To measure the agreement between domain gradients, we define the following function:
\begin{equation}
    \Phi(g^{(1)},...,g^{(N)})_{k} =
    \begin{cases}
        1, & \text{sgn}(g^{(1)}_k)=...=\text{sgn}(g^{(N)}_k) \\
        0, & \text{otherwise},
    \end{cases}
\end{equation}

\noindent where $\text{sgn}(\cdot)$ is the sign function and $g^{(i)}_k$ denotes the $k$-th component of the gradient associated to the $i$-th source domain. The gradient agreement function $\Phi$ checks element-wise if the signs of the gradient components match. When all components have the same sign for a given $k$, it returns 1; if there is any difference, it returns 0. In other words, $\Phi : \mathbb{R}^n \times ... \times \mathbb{R}^n \rightarrow \{0,1\}^n$ takes a set of $N$ gradient vectors as input and returns a new binary vector of the same size $n$. Note that the total size of the gradient vectors will be given by the number of neural parameters, i.e. $n = |\theta|$. $\Phi$ acts as a component-by-component indicator function, where 1 indicates agreement and 0 indicates conflict. In terms of computational complexity, it follows that $\Phi$ is applied to the $N$ domain gradients, so it scales with the number of training domains. The number of domains is expected not to be large ($N = 3$ in our case), thus avoiding potential issues of computational requirements for large $N$ values.

The next step is to define the value of each component for the consensus gradient $g^{*}$, which will be used to update the model parameters $\theta$. For this purpose, we adopt two different rules depending on the value returned by $\Phi_k$. The value of the $k$-component of $g^{*}$ is defined as follows:

\begin{equation}
    g^{*}_{k} =
    \begin{cases}
        \sum_{i=1}^{N} g^{(i)}_k, & \text{if }\Phi_k=1 \\
        0, &  \text{if }\Phi_k=0.
    \end{cases}
    \label{eq:agrsum}
\end{equation}

\noindent Note that $\Phi_k=1$ indicates that gradient component $k$ agrees along all the domains, so we proceed to sum the corresponding values. In contrast, when there is no agreement ($\Phi_k=0$), we resolve the conflict by setting it to zero. In this way, we avoid updating neural weights when there is no consensus, reducing the amount of harmful gradient interference between domains.
A similar approach was derived in \cite{parascandolo2020learning} following the notion of \textit{invariances}, to improve consistency across different domains as the opposite of averaging gradients\footnote{Revised October 2021.}.\\

\noindent \textbf{Agr-Rand consensus strategy.} We also propose an alternative strategy that uses the same approach that Arg-Sum to detect conflicting gradient components via the agreement function $\Phi$, but differs in how conflicts are solved. As before, when there is total agreement (i.e. when $\Phi_k=1$) we sum the gradient components. However, instead of setting to 0 the conflicting components when they do not agree (i.e. when $\Phi_k=0$), Agr-Rand assigns a random value to the consensus gradient by sampling from a normal distribution as follows:

\begin{equation}
    g^{*}_{k} =
    \begin{cases}
        \sum_{i=1}^{N} g^{(i)}_k, & \text{if }\Phi_k=1 \\
        g^{*}_{k} \sim \mathcal{N}(0,\sigma^2), &  \text{if }\Phi_k=0.
    \end{cases}
    \label{eq:agrrand}
\end{equation}

The rationale behind this approach is that zeroing out the conflicting components may lead to dead weights that are never modified during training. Thus, by assigning random values centered at $0$, we may avoid this effect. Note that the Gaussian distribution has zero mean and its variance is given by $\sigma^2$.  We keep all the gradient components within the same range by defining $\sigma^2$ based on the mean absolute value of those components of $g^{*}$ that agree. In other words, if we denote with $\mathcal{A}$ the set of indices $p$ such that $\Phi_p=1$, then $\sigma^2=(\frac{1}{|\mathcal{A}|} \sum_{p \in \mathcal{A}} |\text{g}^{*}_{p} |)^2$. In this way, we assign positive or negative random values sampled from a controlled range.

\subsection{Baseline models}
We compared the proposed methods with a baseline procedure following the standard approach (Deep-All) and the original MTL gradient surgery method (PCGrad), that we adapted to the DG context. Deep-All uses standard mini-batch gradient descent, where the mini-batches are built by randomly sampling images from all the source domains. 

PCGrad \cite{yu2020gradient} takes a task $i$ and computes the cosine similarity between the gradient $g^{(i)}$ and the gradient $g^{(j)}$ of a different task $j$; if the value is negative, it proceeds to replace $g^{(i)}$ by projecting it onto the normal plane of $g^{(j)}$, that is: 

\begin{equation}
    g^{(i)}=g^{(i)}-\frac{\langle g^{(i)},  g^{(j)} \rangle}{\norm{g^{(j)}}^2} g^{(j)}.
    \label{eq:pcgrad}
\end{equation}

\noindent This process is repeated across all other tasks $j \neq i$ sampled in random order. Finally, all the projected task-gradients $g^{(i)}$ are summed to obtain the final gradient. We transfer this idea to the DG context by considering domain gradients instead of task gradients.

We also included four DG state-of-the-art (SOTA) methods in the comparison: Invariant Risk Minimization (IRM) \cite{arjovsky2019invariant}, Meta-Learning Domain Generalization (MLDG) \cite{li2018learning}, Inter-domain Mixup (Mixup) \cite{yan2020improve} and Group Distributionally Robust Optimization (DRO) \cite{sagawa2019distributionally}. For these methods, we adapted the available implementations from \cite{gulrajani2020search} to our framework. 

\section{Experiments and results}

\subsection{Dataset details}

We evaluated our method on three well-known datasets for multi-domain image classification: PACS \cite{li2017deeper}, VLCS \cite{fang2013unbiased} and Office-Home \cite{venkateswara2017deep}. PACS includes 9,991 images of 4 domains: Art (A), Cartoon (C), Photo (P) and Sketch (S); and 7 classes. VLCS contains 10,729 photographic images of 4 domains: Caltech101 (C), LabelMe (L), SUN09 (S) and VOC2007 (V); organized into 5 classes. Office-Home contains 15,588 images of everyday objects organized into 4 domains: Art (A), Clipart (C), Product (P) and Real-World (R); and 65 classes. Figure \ref{fig:datasets} displays some examples of these datasets. PACS and Office-Home are more challenging than VLCS as they provide non-photographic visual domains (such as paintings and sketches), resulting in a more pronounced domain change. Due to the fact that all images in PACS are 227x227 and in VLCS and Office-Home they have different sizes, we resized all images in VLCS and Office-Home into 227x227 so that the image size is consistent across all datasets.

In order to measure the generalization performance of our method, we adopted the leave-one-domain-out strategy, i.e. holding one domain out for testing and using the remaining domains for training. For all datasets, we randomly split each domain into training (70\%), validation (10\%) and testing (20\%) subsets. Note that the images used to construct the training and validation sets will come from multiple source domains, different to that used for testing. 
For testing, we selected the model that achieves the highest accuracy on the validation set and evaluated it on the test subset of the held-out domain.
\begin{figure*}
\begin{center}
    \includegraphics[width=\textwidth]{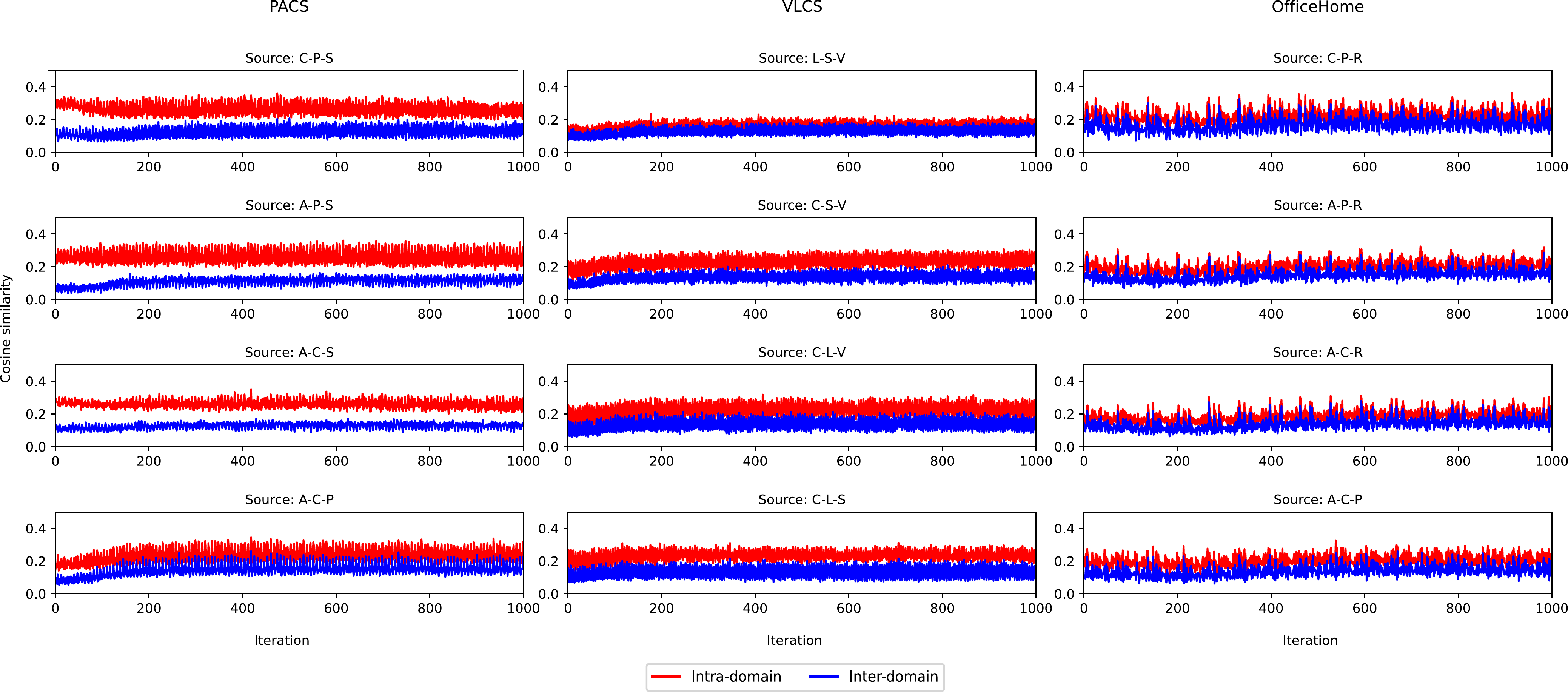}
\end{center}
   \caption{Average gradient cosine similarity within and between domains for the PACS, VLCS and Home-Office datasets for a standard training procedure. Each plot represents a different combination of source domains used for training. We observe that gradients computed for images of the same domain (intra-domain, in red) exhibit higher cosine similarity than images from different domains (inter-domain, in blue). This experiment supports our hypothesis about conflicting gradients emerging in multi-domain scenarios.}
\label{fig:cosine_sim}
\end{figure*}
\subsection{Implementation details}

\noindent \textbf{Network architecture:} Following previous works \cite{li2017deeper, dou2019domain, carlucci2019domain}, we chose a well-known CNN architecture for image classification and then finetuned the network on source domains. For all methods, we used an AlexNet \cite{krizhevsky2012imagenet} pretrained on ImageNet \cite{russakovsky2015imagenet} and reshaped the last fully connected (FC) layer to have the same number of outputs as the number of classes in the respective datasets (7 PACS, 5 VLCS and 65 Office-Home).
Note that here we chose a relatively simple architecture since it was faster to train and served as a proof-of-concept to analyze the impact of gradient surgery methods on domain generalization. Thus, we focus on the relative improvement achieved by gradient surgery with respect to the baseline Deep-All model. However, as our method is agnostic to model architecture, more complex networks producing higher baseline results (like ResNet \cite{he2016deep} or Inception \cite{szegedy2015going}) could be used instead.\\

\noindent \textbf{Implementation:} All experiments were implemented in PyTorch \cite{paszke2019pytorch} and run in a machine with CPU Intel Core i7-8700, 32GB RAM and NVidia Titan Xp GPU. We trained all models with cross-entropy loss function during 1000 iterations or up to convergence, validating every 20 steps. At every training iteration, we randomly sampled a batch of size 128 from each source domain. For optimization, we used the Adam optimizer \cite{kingma2014adam} and as a regularization technique we employed weight decay. The learning rate and the regularization parameter $\lambda$ were adjusted by grid search using the validation set, and the resulting values were 1e-5 and 5e-5, respectively, for all methods and datasets.\footnote{Our source code is publicly available at \url{https://github.com/lucasmansilla/DGvGS}.}

\subsection{Gradient characterization for multiple domains}
Our working hypothesis is that when training with multiple domains, conflicting gradients within each mini-batch contain information specific to the individual domains which is irrelevant to the others, including the test domain. To shed light on this matter, we designed a study to characterize the gradients emerging while training with multiple domains. 
We measure how similar the gradients are within and between domains using cosine similarity. In order to avoid a possible interference given by the different classes, we decided to use data from the same class at every training iteration, so that the only source of differences is the domain. During training, we sample a mini-batch from each source domain, taking care that in every iteration we only select samples from a given class. Gradients for the loss function are computed individually, for each sample of the mini-batch. The alignment between gradients is then measured by cosine similarity, considering pairs of gradients from the same domain and from different domains. Figure \ref{fig:cosine_sim} shows the average cosine similarity within and between domains for the PACS, VLCS and Home-Office datasets. Note that in all cases the gradients tend to exhibit higher similarity within domains than between domains. This confirms that pairs of inter-domain gradients carry more conflicting information than the intra-domain ones. In the next experiment, we will show that reducing such interference by encouraging gradient agreement tends to improve generalization in unseen domains. 
\begin{figure*}
\begin{center}
    \includegraphics[width=\textwidth]{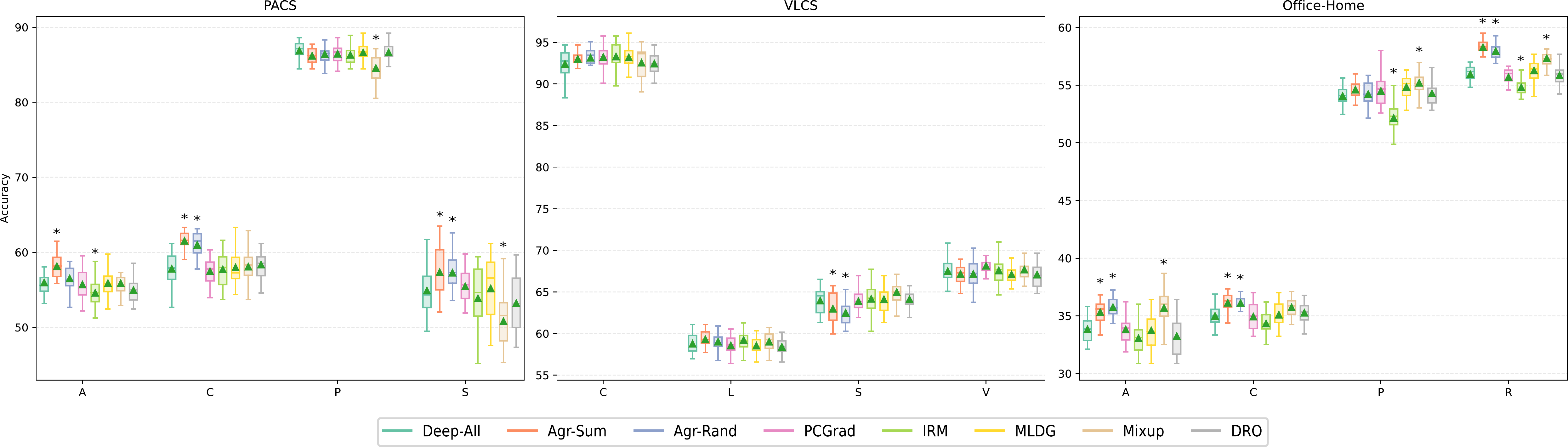}
\end{center}
   \caption{Accuracy of leave-one-domain-out evaluation on PACS, VLCS and Office-Home datasets. The target domain (unseen during training) is specified below each group of boxplots. Each boxplot represents 20 independent runs; the box shows the values from the lower to upper quartile, the line is the median, the green triangle is the mean and the whiskers show the minimum and maximum values. The asterisk (*) above a boxplot of a method indicates that differences between the means of that method and Deep-All are significant at a 0.05 level according to a paired Wilcoxon test.}
\label{fig:boxplots}
\end{figure*}

\subsection{Evaluating the impact of gradient surgery for domain generalization}
In this experiment, we evaluate the impact of the proposed gradient surgery strategies for improved domain generalization. To account for possible differences due to network initialization, we performed 20 independent runs for each combination of dataset, method and held-out domain. For each of them, we report the average accuracy on the test subset of the held-out domain. Results are shown in Figure  \ref{fig:boxplots} and Table \ref{tab:accuracy}. We evaluated the statistical difference between the mean accuracy reported for Deep-All (baseline) and the other methods: the alternative gradient surgery approaches (Agr-Sum, Agr-Rand and PCGrad) and the SOTA methods (IRM, MLDG, Mixup and DRO). 
We used the paired Wilcoxon test for statistical significance in terms of mean difference (with a significance level of 0.05). 

Figure \ref{fig:boxplots} shows the accuracy of the different methods on PACS, VLCS and Office-Home datasets. In PACS, we can observe that Agr-Sum and Agr-Rand significantly outperform the Deep-All baseline in 3 of the 4 target domains (Art-Painting, Cartoon and Sketch). Similarly, in Office-Home, the aforementioned methods improve generalization performance in 3 of the 4 target domains (Art, Clipart and Real-World). No improvements in performance are observed in VLCS that favor the use of a particular method over a different one. This may be due to the fact that the overall accuracy in VLCS is already higher than in PACS and Office-Home. This fact may leave smaller room for improvement than in the other cases, making the differences less significant. Moreover, while in VLCS all domains correspond to photographs, in PACS and Office-Home we also find art-painting, cartoons, cliparts and sketches (see Figure \ref{fig:datasets} for visual examples). The proposed gradient agreement strategies seem to be more useful in such multi-modal scenarios. This is coherent with observations made in a previous work \cite{li2017deeper}, which reports larger Kullback-Leibler divergence for inter-domain features from PACS than from VLCS, and larger improvements for PACS than VLCS with respect to a Deep-All baseline. 

The mean accuracy and standard deviation across all domains for each method are reported in Table \ref{tab:accuracy}. From these results, we can see that Agr-Sum, Agr-Rand and PCGrad perform better than Deep-All and the SOTA methods in 8 of the 12 evaluations (6 Agr-Sum, 1 Agr-Rand and 1 PCGrad). Moreover, within each dataset, Agr-Sum and Agr-Rand outperform Deep-All and the SOTA methods on average in PACS and Office-Home, and they are competitive in VLCS. Overall, we observe that significant improvements favor the use of Agr-Sum as agreement strategy, specially in scenarios with pronounced domain shift. Moreover, the relative improvement achieved by gradient surgery with respect to the baseline Deep-All model is actually on par with that reported in previous works \cite{li2017deeper, dou2019domain, carlucci2019domain}.

When comparing the proposed gradient surgery strategies (Agr-Sum and Agr-Rand) with PCGrad in the context of domain generalization, we observe that PCGrad tends to replicate the results of Deep-All in most of the cases. In other words, gradient surgery fails to significantly boost performance in this case. However, when used in MTL settings, PCGrad had shown to be effective, as discussed in \cite{yu2020gradient}. It remains to be elucidated why it is the case that PCGrad does not help in our study. One possible reason is that more subtle differences in terms of gradient conflicts emerge in multi-domain scenarios compared to multi-task cases. The strategy followed by Agr-Sum and Agr-Rand, i.e. zeroing out or assigning random values to the conflicting components, seems to be more aggressive than projecting onto the normal component of the other tasks. Thus, PCGrad may be enough to harmonize gradients and make a difference in the context of MTL, but not in case of multi-domain scenarios. However, verifying this hypothesis will require to implement an experimental setting that allows comparison between multi-domain and multi-task learning under similar conditions. Further studies are required to confirm this presumptions, which are left as future work.

\begin{table*}
\begin{center}
\begin{adjustbox}{width=0.95\textwidth}
\begin{tabular}{*{11}{c}}
\toprule
 & & &
\multicolumn{8}{c}{Method} \\ 
\cmidrule(l){4-11}
& \multicolumn{2}{c}{Training schedule} &
\multicolumn{1}{c}{Baseline} &
\multicolumn{3}{c}{Gradient surgery} &
\multicolumn{4}{c}{SOTA} \\ 
\cmidrule(lr){2-3} \cmidrule(lr){4-4} \cmidrule(lr){5-7} \cmidrule(l){8-11}
Dataset & Source & Target & Deep-All & Agr-Sum & Agr-Rand & PCGrad & IRM & MLDG & Mixup & DRO \\
\toprule
\multirow{5}{*}{PACS}
& C,P,S & A & 55.98 (1.75) & \textbf{58.13 (1.65)}* & 56.51 (1.48) & 55.70 (2.02) & 54.59 (1.98)* & 55.88 (1.92) & 55.88 (1.65) & 54.96 (1.55) \\
& A,P,S & C & 57.80 (2.21) & \textbf{61.52 (1.21)}* & 60.99 (1.55)* & 57.47 (1.79) & 57.72 (2.37) & 57.99 (2.14) & 58.08 (1.95) & 58.36 (2.32) \\
& A,C,S & P & \textbf{86.87 (1.22)} & 86.18 (1.09) & 86.41 (1.25) & 86.47 (1.25) & 86.30 (1.23) & 86.63 (1.14) & 84.55 (1.76)* & 86.63 (1.03) \\
& A,C,P & S & 54.90 (3.28) & \textbf{57.35 (3.29)}* & 57.27 (2.97)* & 55.46 (2.91) & 53.86 (4.22) & 55.18 (4.24) & 50.81 (4.08)* & 53.21 (3.70) \\
\cmidrule{2-11}
& & Avg. & 63.89 & 65.80 & 65.30 & 63.77 & 63.12 & 63.92 & 62.33 & 63.29 \\
\midrule
\multirow{5}{*}{VLCS}
& L,S,V & C & 92.40 (1.81) & 93.00 (0.94) & 93.14 (1.28) & 93.23 (1.50) & \textbf{93.29} (1.61) & 93.18 (1.45) & 92.54 (1.96) & 92.44 (1.23) \\
& C,S,V & L & 58.78 (1.07) & \textbf{59.30 (1.07)} & 59.02 (1.12) & 58.56 (1.17) & 59.22 (1.49) & 58.55 (1.11) & 59.02 (1.12) & 58.40 (1.04) \\
& C,L,V & S & 63.96 (1.63) & 62.98 (1.85)* & 62.50 (1.68)* & 63.89 (1.25) & 64.16 (1.87) & 64.11 (1.70) & \textbf{64.98 (1.40)} & 64.11 (1.17) \\
& C,L,S & V & 67.49 (1.49) & 67.15 (1.10) & 67.15 (1.58) & \textbf{68.14 (0.97)} & 67.57 (1.41) & 67.10 (1.07) & 67.68 (1.38) & 67.08 (1.53) \\
\cmidrule{2-11}
& & Avg. & 70.66 & 70.61 & 70.45 & 70.96 & 71.06 & 70.74 & 71.06 & 70.51 \\
\midrule
\multirow{5}{*}{Office-Home}
& C,P,R & A & 33.84 (1.14) & 35.32 (1.02)* & \textbf{35.75 (0.86)}* & 33.82 (1.12) & 33.07 (1.28) & 33.73 (1.55) & 35.69 (1.51)* & 33.25 (1.55) \\
& A,P,R & C & 34.99 (1.37) & \textbf{36.13 (0.88)}* & 36.12 (0.88)* & 34.94 (1.18) & 34.34 (1.07) & 35.10 (1.08) & 35.74 (0.87) & 35.27 (0.95) \\
& A,C,R & P & 54.06 (0.95) & 54.22 (1.06) & 54.22 (1.06) & 54.49 (1.30) & 52.16 (1.26)* & 54.85 (1.03) & \textbf{55.20 (1.02)}* & 54.28 (0.97) \\
& A,C,P & R & 55.95 (0.89) & \textbf{58.29 (0.78)}* & 57.95 (0.70)* & 55.71 (0.84) & 54.81 (0.89)* & 56.27 (0.98) & 57.33 (0.86)* & 55.84 (0.88) \\
\cmidrule{2-11}
& & Avg. & 44.71 & 46.09 & 46.01 & 44.74 & 43.59 & 44.99 & 45.99 & 44.66 \\
\bottomrule
\end{tabular}
\end{adjustbox}
\end{center}
    \caption{Mean accuracy and standard deviation of leave-one-domain-out evaluation on PACS, VLCS and Office-Home datasets. For each dataset, we also report the average accuracy of the different methods over all target domains. The method that achieves the highest accuracy on a given target domain is indicated in bold in each row. The asterisk (*) indicates that the difference with respect to Deep-All is statistically significant.}
\label{tab:accuracy}
\end{table*}

\begin{figure}
\begin{center}
    \includegraphics[width=\columnwidth]{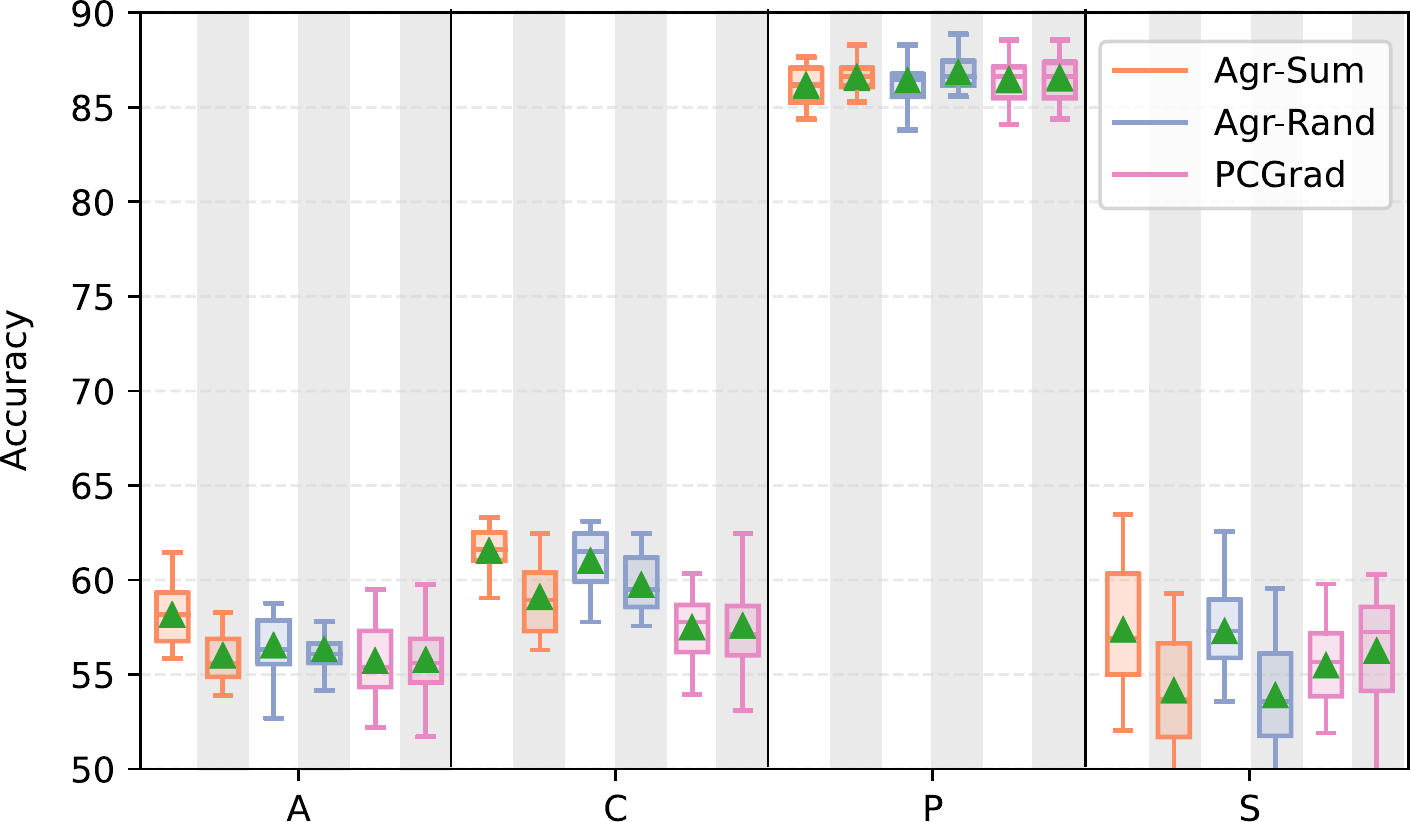}
\end{center}
   \caption{Control experiment on PACS comparing gradient surgery using multi-domain batches (white) vs single-domain batches (grey shaded).}
\label{fig:control}
\end{figure}

We also performed a control experiment to analyze whether the improvement obtained by our gradient surgery was due to the inter-domain gradient agreement, or it just was a simple regularization effect coming from the gradient surgery itself. To this end, we evaluated the effect of training a model with gradient surgery on multiple batches where each one is sampled from a different domain (\textit{multi-domain}), compared to training on multiple batches sampled from a single domain randomly chosen at each training iteration (\textit{single-domain}). Note that in both cases we used 3 domains for training, and the difference is that during a single gradient descent iteration the gradients $g^{(i)}$ in multi-domain come from different domains, while they come from the same one in single-domain. Figure \ref{fig:control} shows the average accuracy of 20 independent runs on PACS for Agr-Sum, Agr-Rand and PCGrad using multi-domain and single-domain batches. From these results, we can notice that there are differences in accuracy favoring Agr-Sum and Agr-Rand in 3 out of 4 target domains when training with multi-domain batches. This shows that gradient agreement contributes to effectively improve the generalization performance by encouraging the inter-domain gradient agreement in the batches. 

\section{Conclusions}
In this work we studied the implications of multi-domain gradient interference in domain generalization, and proposed alternative gradient surgery strategies to mitigate their negative effect. Our characterization of intra and inter-domain gradients confirmed the initial hypothesis that pairs of inter-domain gradients carry more conflicting information than the intra-domain gradients. Experiments on three multi-domain datasets showed that gradient agreement strategies are useful in reducing inter-domain interference and tend to improve generalization in unseen domains. Our comparative study with the Deep-All baseline, the PCGrad agreement strategy and the SOTA methods shows that the proposed Agr-Sum method outperforms the other strategies in most scenarios. Such improvement is more clear in cases where domain shift leads to poor performance of the baseline model. This is the case of target domains A, C and S in PACS or A and C in Office-Home, which present a low baseline performance that is significantly improved when using Agr-Sum. 

The proposed gradient surgery methods are agnostic to model architecture and do not augment the number of hyper-parameters. In the future, we plan to explore their impact when training more complex deep neural architectures which should lead to higher performance. 


\section*{Acknowledgments}

 We thank Siddhartha Chandra for the useful comments and discussion. The authors gratefully acknowledge NVIDIA Corporation with the donation of the GPUs used for this research, and the support of UNL (CAID-0620190100145LI, CAID-50220140100084LI) and ANPCyT (PICT). This work was supported by Argentina's National Scientific and Technical Research Council (\mbox{CONICET}), who covered all researchers salaries.

{\small
\bibliographystyle{ieee_fullname}
\bibliography{references}
}

\end{document}